\documentclass{article}

\usepackage[preprint, nonatbib]{neurips_2024}

\usepackage[utf8]{inputenc} 
\usepackage[T1]{fontenc}    
\usepackage{hyperref}       
\usepackage{url}            
\usepackage{booktabs}       
\usepackage{amsfonts}       
\usepackage{nicefrac}       
\usepackage{microtype}      
\usepackage{xcolor}         
\usepackage{amsmath}      
\usepackage{graphicx}
\usepackage{array}
\usepackage{float} 
\usepackage{caption} 
\usepackage{subcaption} 
\usepackage{placeins}
\usepackage{makecell}
\usepackage{lscape}
\usepackage{wrapfig}
\title{Deep Learning Methods for the Noniterative Conditional Expectation G-Formula for Causal Inference from Complex Observational Data}

\author{%
  Sophia M. Rein\thanks{Equal contribution.} \\
  CAUSALab \\
  Harvard T.H. Chan School of Public Health \\
  Boston, MA 02115 \\
  \texttt{srein@hsph.harvard.edu} \\
  \And
  Jing Li\footnotemark[1] \\
  CAUSALab  \\
  Harvard T.H. Chan School of Public Health \\
  Boston, MA 02115 \\
  \texttt{jing\_li@hsph.harvard.edu} \\
  \And
  Miguel Hernan\thanks{Co-senior authors.} \\
  CAUSALab  \\
  Harvard T.H. Chan School of Public Health \\
  Boston, MA 02115 \\
  \texttt{mhernan@hsph.harvard.edu} \\
  \And
  Andrew Beam\footnotemark[2] \\
  CAUSALab  \\
  Harvard T.H. Chan School of Public Health \\
  Boston, MA 02115 \\
  \texttt{Andrew\_Beam@hms.harvard.edu} \\
}

\begin{document}

\maketitle

\begin{abstract}
The g-formula can be used to estimate causal effects of sustained treatment strategies using observational data under the identifying assumptions of consistency, positivity, and exchangeability. The non-iterative conditional expectation (NICE) estimator of the g-formula also requires correct estimation of the conditional distribution of the time-varying treatment, confounders, and outcome. Parametric models, which have been traditionally used for this purpose, are subject to model misspecification, which may result in biased causal estimates. Here, we propose a unified deep learning framework for the NICE g-formula estimator that uses multitask recurrent neural networks for estimation of the joint conditional distributions. Using simulated data, we evaluated our model's bias and compared it with that of the parametric g-formula estimator. We found lower bias in the estimates of the causal effect of sustained treatment strategies on a survival outcome when using the deep learning estimator compared with the parametric NICE estimator in settings with simple and complex temporal dependencies between covariates. These findings suggest that our Deep Learning g-formula estimator may be less sensitive to model misspecification than the classical parametric NICE estimator when estimating the causal effect of sustained treatment strategies from complex observational data.
\end{abstract}

\section{Introduction}

The g-formula (or g-computation algorithm formula) identifies the causal effect of sustained treatment strategies under the conditions of consistency, positivity, and exchangeability\cite{Robins}. Unlike previously proposed methods for causal inference, the g-formula is valid even in the presence of treatment-confounder feedback, i.e., when time-varying confounders are affected by past treatment as commonly occurs in medical settings \cite{Hernan}.  \\

A common representation of the g-formula is a non-iterative conditional expectation (NICE) of the outcome weighted by the joint density of covariate history\cite{Wen}. The NICE g-formula estimator has 2 stages: (i) estimation of the conditional distributions of time-varying confounders, treatments, and outcome, and (ii) calculation of an integral (or sum) over all possible covariate histories, which is usually approximated via Monte Carlo simulation. \\

In realistic settings that require adjustment for a high-dimensional set of confounders, nonparametric estimation of the components of the NICE g-formula is usually not possible. Therefore, in real-world applications\cite{Lodi}\cite{Taubman}\cite{Dickerman}\cite{Jain}, the NICE g-formula estimator has been based on parametric models such as generalized linear models (GLMs).

Because correct specification of parametric models may be difficult \cite{Robins2}\cite{McGrath}, an alternative is the use of more flexible deep neural networks (aka "deep learning") that do not require an explicit specification of the functional form of covariates. Specifically, recurrent neural networks, such as long short-term memory (LSTM) models\cite{Hochreiter}, were developed to model longitudinal data, and thus are well suited to model evolving covariate trajectories with potentially long-range dependencies between them. Therefore, LSTMs can be used to jointly predict all covariate trajectories in stage (i) of the NICE g-formula estimator. A previous study \cite{Li} found that g-formula estimates based on LSTMs were less biased than those based on GLMs. However, this study implemented GLMs that are too simplistic for most practical applications. Therefore, it is unclear whether LSTMs would similarly outperform the traditional parametric approach when compared with sufficiently complex and richly parameterized GLMs.\\

Here we describe a deep learning (DL) NICE g-formula estimator that uses LSTMs to estimate the joint distribution of covariates and outcomes. Using simulated data that mimics real-world data, we compare the bias of g-formula estimates between this DL-NICE g-formula estimator and the parametric NICE g-formula estimator when targeting the causal effect of sustained treatment strategies on a survival outcome.

\section{Methods}

\subsection{Simulated data}

We simulated a cohort of people with HIV who started antiretroviral treatment for the first time. We used this simulated dataset to mimic a recent study that estimated the effect of integrase inhibitors (INSTI; a class of antiretrovirals) on the risk of cardiovascular events\cite{Rein} in the HIV-CAUSAL Collaboration, a consortium of observational studies from Europe and North America\cite{Lodi}.\\

The data simulation procedures are described in the supplementary material (see Section~\ref{sec:sim_design}). Briefly, we simulated data for each individual at months $k =0, 1, 2, \dots, K$, where $k = 0$ is baseline and $K = 60$ is the end of the follow-up. For each individual and month, we simulated a time-varying indicator $A_k$ for initiation of an antiretroviral combination regimen containing an INSTI drug during $k$, a vector of time-varying confounders $L_k$ measured at the start of month $k$, and a time-varying indicator $Y_{k}$  of experiencing a cardiovascular event (myocardial infarction, stroke, or invasive cardiovascular procedure) during month $k$. No individual was lost to follow-up. \\

We simulated datasets with two different dependencies between variables across time. First, we used a 'simple' time-dependency between covariates: the value of every variable at $k$ does not depend on covariate values before time $k-1$. Second, we used a 'complex' time-dependency between covariates: the value of every variable at $k$ depends on a function of the entire history of all covariates through $k$. For each type of dependency, we simulated datasets of 1,000 and 10,000 individuals. 

\subsection{The g-formula and the plug-in NICE estimator}

The risk by $K$ had all individuals received treatment according to an intervention distribution $f^{int}(a_k|Y_k=0,\bar{l}_{k},\bar{a}_{k-1})$ is identified by the g-formula (see equation \ref{full_gform} in supplementary material) under exchangeability, positivity and consistency\cite{Robins}. For a deterministic treatment strategy (i.e. $f^{int}(a_k|Y_k=0,\bar{l}_{k},\bar{a}_{k-1})$ either equals 0 or 1 for all $(\bar{l}_k,\bar{a}_{k-1})$) the joint density g-formula equals $E(Y_K^g)$ and is given by:
\begin{equation} \label{det_gform}
\begin{split}
\sum_{\forall{\bar{l}_{K-1}}} \sum_{k=1}^{K} P(Y_k=1|Y_{k-1}=0,\bar{L}_{k-1} =\bar{l}_{k-1},\bar{A}_{k-1}=\bar{a}_{k-1}^g) \times \\
\prod_{s=0}^{k-1} P(Y_s=0||Y_{s-1}=0,\bar{L}_{s-1} =
\bar{l}_{s-1},\bar{A}_{s-1}=\bar{a}_{s-1}^g) f(l_s|Y_s=0,\bar{l}_{s-1},\bar{a}_{s-1}^g),
\end{split}
\end{equation}

where $\bar{X}_k=(X_0,...,X_k)$ denotes the history of random variable $X$, through $k$. By definition, $Y_0 = 0$ and $\bar{L}_{-1}=\bar{A}_{-1}=\emptyset$.
\\

The plug-in parametric estimator of the NICE g-formula has often been implemented with the following steps: \\
1. Fit models for each component in Expression \ref{det_gform}. Specifically,\\
(a) Fit a pooled (over persons and time) model for the conditional distribution of each confounder in $L_k$ at time $k$ as a function of k, past treatment, and confounder history based on those who are alive at $k$.\\
(b) Fit a pooled logistic regression model for the probability of experiencing the outcome of interest by time $k+1$ as a function of $k$, past treatment, and confounder history based on those who are alive at $k$. \\
2. Approximate the integral in expression \ref{det_gform} via Monte Carlo simulation. For each simulation sample $m=1,...,n$, baseline confounders at $k=0$ are sampled from observed values, and treatment at $k=0$ is assigned according to the intervention rule. For $k>0$:\\
(a) Simulate confounders from the fitted models in Step 1(a) using previously simulated confounders and assigned treatment values through time $k-1$. Assign treatment according to the intervention based on simulated confounders and assigned treatment values through time $k-1$.\\
(b) Compute the discrete-time hazard of the outcome at time $k+1$ from the fitted model in Step 1(b) using previously simulated confounders and assigned treatment values through time $k$.\\
3. Calculate the average cumulative probability of failure by time $K$ over all generated intervention histories.

\subsection{Deep learning NICE algorithm}

The DL-NICE g-formula algorithm replaces the parametric models for covariates and outcomes by a single multitask recurrent neural network for joint prediction of all covariates and a recurrent neural network for the outcome. The recurrent neural networks do not require specification of the functional form of the models (i.e., the relationships between covariates), which may be complex and unknown. The DL-NICE algorithm has the following steps depicted in Figure \ref{fig1}: 

\begin{wrapfigure}{r}{0.65\textwidth} 
  \centering
  \includegraphics[width=0.63\textwidth]{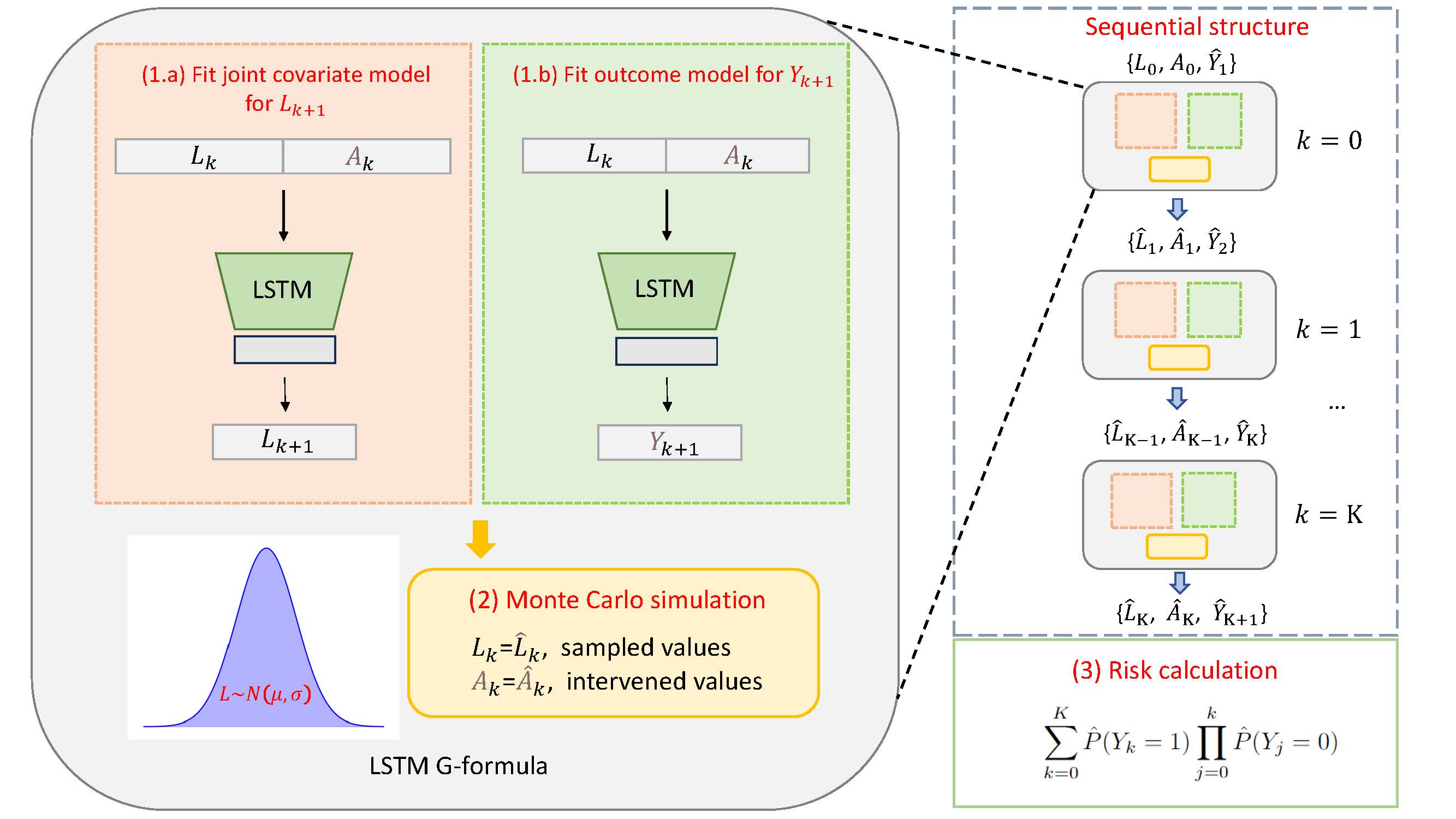} 
  \caption{Deep Learning NICE g-formula framework}
  \label{fig1}
\end{wrapfigure}

\emph{Step 1: Representation learning by the recurrent neural network}
(a) Fit a recurrent neural network (we fit a LSTM in the present study but other recurrent neural networks could be considered) to jointly estimate the conditional distributions of the confounders. The model receives as input all covariate and treatment values up to and including at time $k-1$ (i.e. the entire history of treatment and confounders).
(b) Fit a recurrent neural network (here, LSTM) to estimate the probability of experiencing the outcome of interest at each time $k+1$ based on those who are alive at $k$. The model receives as input all covariate and treatment values up to and including at time $k$; i.e. the entire history of treatment and confounders).

\emph{Step 2: Monte Carlo simulation of covariate trajectories and computation of outcome hazards}
Monte Carlo simulation is performed for $n$ number of times based on the intervention of interest. The covariate values are sequentially simulated by assuming a normal distribution for the mean $\mu$ of the covariate distributions estimated by the LSTM. The procedure for computation of the hazards is the same as in the classic parametric g-formula algorithm described above.

\emph{Step 3: Calculate the cumulative incidence of the outcome}
Calculate the average cumulative probability of failure by time $K$ over all generated intervention histories as is done in the classical g-formula algorithm.

\subsection{Analysis}
We investigated the 5-year risk of cardiovascular events under two deterministic static treatment strategies (see section~\ref{sec:treat_strat} in the supplementary material for a formal definition of treatment strategies): (1) “initiate ART containing INSTI at time 0 and continue to treat with this combination during the study” (or “always treat with INSTI”), and (2) “initiate ART not containing INSTI at time 0 and continue with this strategy throughout the study” (or “never treat with INSTI”). We first estimated the absolute risk under no intervention on treatment (the "natural course") and under the "always treat" and "never treat" strategies and then estimated the causal contrast (risk ratio and risk difference) comparing the "always treat" and "never treat" strategies using first the classical parametric g-formula estimator and then our DL-NICE estimator. We then evaluated the absolute bias in the estimated risks over time under each strategy and the bias in the estimates of the causal effect (risk ratio and risk difference) under both methods and averaged the bias across all time points. We obtained the 'ground truth' risks from simulated datasets with 1 million observations for the 'simple' and 'complex' time dependency scenario. Let \(\hat{R}^{\text{method}}_k\) denote the estimated risk at time \(k\) obtained using either the DL-NICE estimator or the classical parametric g-formula estimator, and let \(R^{\text{true}}_k\) represent the 'ground truth' risk at time \(k\). The bias in the estimated risk at time \(k\) is then defined as:
\[
\text{Bias}(\hat{R}^{\text{method}}_k) = \hat{R}^{\text{method}}_k - R^{\text{true}}_k.
\]
Similarly, for the causal effect estimates, the risk ratio (RR) and risk difference (RD), the biases were calculated as the differences between the estimates obtained from the methods and the corresponding ground truth values. For the risk ratio, the bias is defined as:
\[
\text{Bias}(\hat{\text{RR}}^{\text{method}}_k) = \hat{\text{RR}}^{\text{method}}_k - \text{RR}^{\text{true}}_k.
\]
For the risk difference, the bias is defined as:
\[
\text{Bias}(\hat{\text{RD}}^{\text{method}}_k) = \hat{\text{RD}}^{\text{method}}_k - \text{RD}^{\text{true}}_k.
\]
We conducted these analyses in both the dataset with the 'simple' and 'complex' time dependency scenario and for both sample sizes (1,000 and 10,000). To optimize hyperparameter values for each LSTM, we used the 'hyperopt' Python package with 50 trials for each search. The hyperparameter search strategy and settings that we used for the LSTMs are shown in the supplementary material in section ~\ref{sec:hyper_param}. \\
For the classic parametric g-formula, we fit different parametric models. For the simple time dependency scenario, we fit parametric models for the covariates and outcome that were equal to the data generating process (i.e. each covariate depends on covariate values at time $k_{-1}$; see supplemetary material~\ref{sec:sim_design}) except for not including the unmeasured covariate $U$. 
For the complex time dependency scenario, we fit two parametric models that differed in the complexity of covariate histories included:\\
\emph{1. Least flexible}: We fit models for $L_k$ and $Y_k$ that include terms for the lagged value of $L_k$ and $A_k$ (i.e. the values of $L_{k-1}$ and $A_{k-1}$).\\
\emph{2. Moderately flexible}: We fit models for $L_k$ and $Y_k$ that include terms for the lagged value as well as the lagged cumulative average of the covariate values from $k=0$ up to $k-1$.

\section{Results}

\subsection{Simple time-dependency scenario}

Figure 2 shows the ground truth and predicted (using both DL-NICE and parametric NICE) risk of the outcome over time under no intervention (the natural course) and suggests that the risk predicted using the DL-NICE estimator is closer to the ground truth risk. Tables 1 and 2 show the absolute bias averaged across all 60 time points in the estimates of the risk under natural course, the 'always treat' and 'never treat' interventions and the bias in the risk ratio and risk difference comparing the always and never treat interventions (see the supplementary material, section ~\ref{sec:risks_simple}, for the risks under intervention, effect estimates and bias over time). In both sample sizes, the bias in the risk estimates was lower when using the DL-NICE method compared to the parametric method. However, when using the DL-NICE method, the bias in the effect estimates (risk difference and risk ratio) was somewhat higher in the smaller dataset with sample size of 1,000, while it was lower compared to the parametric NICE method in the larger dataset. 

\begin{figure}[H]
\centering

\begin{subfigure}{0.49\textwidth}
    \includegraphics[width=\linewidth]{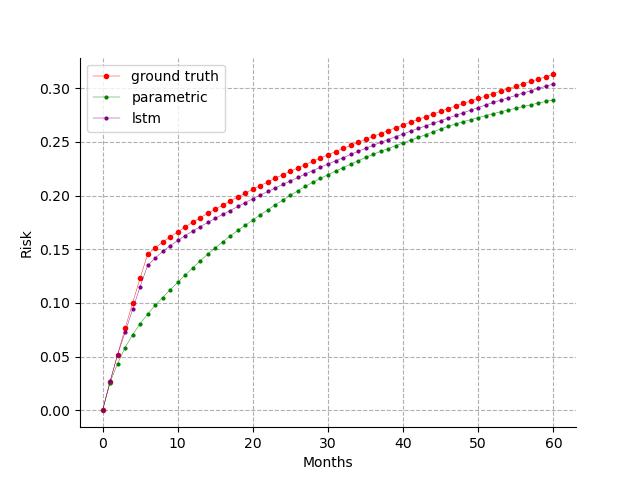} 
    \caption{Sample size = 1,000}
    \label{fig:sfig1}
\end{subfigure}
\hfill
\begin{subfigure}{0.49\textwidth}
    \includegraphics[width=\linewidth]{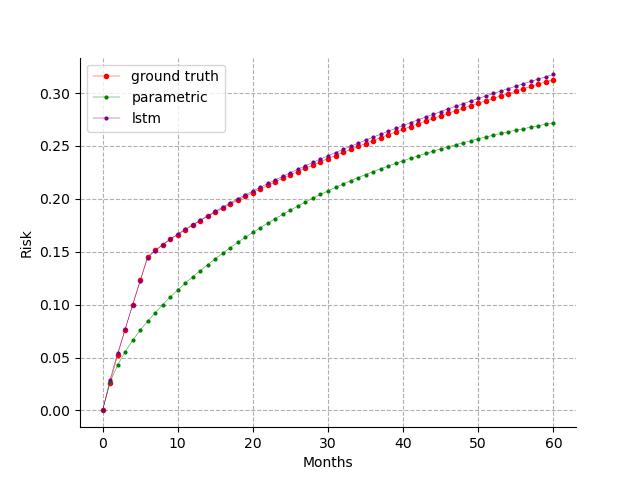} 
    \caption{Sample size = 10,000}
    \label{fig:sfig2}
\end{subfigure}
\caption{Ground truth and estimated risk under the 'natural course' using the parametric and DL-NICE methods on 'simple' time-dependency data}
\end{figure}

\vspace{-2em}

\begin{table}[H]
\centering
\resizebox{\textwidth}{!}{%
\begin{tabular}{|l|c|c|c|c|c|}
\hline
\textbf{Method} & \multicolumn{3}{c|}{\textbf{Bias in risk estimates} (\(\downarrow\) is better)} & \multicolumn{2}{c|}{\textbf{Bias in causal effect estimates} (\(\downarrow\) is better)} \\
\hline
 & \textbf{Natural course} & \textbf{Always treat} & \textbf{Never treat} & \textbf{Risk difference} & \textbf{Risk ratio} \\
\hline
Parametric & 0.024 & 0.036 & 0.014 & \textbf{0.022} & \textbf{0.096} \\
DL-NICE (Ours) & \textbf{0.008} & \textbf{0.029} & \textbf{0.004} & 0.032 & 0.156 \\
\hline
\end{tabular}%
}
\vspace{0.5em} 
\caption{Bias in risk and causal effect estimates (comparing always vs. never treat; averaged over all 60 time points) using the parametric and DL-NICE methods on 'simple' time-dependency data with 1,000 individuals.}
\end{table}

\vspace{-2.2em} 

\begin{table}[H]
\centering
\resizebox{\textwidth}{!}{%
\begin{tabular}{|l|c|c|c|c|c|}
\hline
\textbf{Method} & \multicolumn{3}{c|}{\textbf{Bias in risk estimates} (\(\downarrow\) is better)} & \multicolumn{2}{c|}{\textbf{Bias in causal effect estimates} (\(\downarrow\) is better)} \\
\hline
 & \textbf{Natural course} & \textbf{Always treat} & \textbf{Never treat} & \textbf{Risk difference} & \textbf{Risk ratio} \\
\hline
Parametric & 0.035 & 0.042 & 0.029 & 0.013 & 0.047 \\
DL-NICE (Ours) & \textbf{0.003} & \textbf{0.003} & \textbf{0.004} & \textbf{0.006} & \textbf{0.031} \\
\hline
\end{tabular}%
}
\vspace{0.5em} 
\caption{Bias in risk and causal effect estimates (comparing always vs. never treat; averaged over all 60 time points) using the parametric and DL-NICE methods on 'simple' time-dependency data with 10,000 individuals.}
\end{table}

\subsection{Complex time-dependency scenario}

Figure 3 shows the ground truth and predicted outcome risk over time under the natural course on the complex time-dependency data, again suggesting that the risk predicted using the DL-NICE estimator is closer to the ground truth risk than that predicted using parametric NICE.
Tables 3 and 4 show the absolute bias averaged over all 60 time points in the risk and effect estimates (risk ratio and risk difference) in the datasets with sample sizes of 1,000 and 10,000, respectively (see section ~\ref{sec:risks_complex} in the supplementary material for the estimated risks under the interventions, effect estimates and bias over time). In both sample sizes, the bias in the risk estimates was highest when using the parametric g-formula with the least flexible models for the covariates and the outcome (covariate history includes lagged values only), followed by the parametric g-formula with moderately flexible models (covariate history includes lagged values and the lag of the cumulative average of values over time) and was lowest when using the DL-NICE estimator. However, this difference in bias comparing the parametric and DL-NICE estimator was more pronounced in the larger dataset. The bias in the effect estimates was also lower when using the DL-NICE method for both sample sizes, except for the risk ratio bias in the larger sample size.

\begin{figure}[H]
\centering

\begin{subfigure}{0.49\textwidth}
    \includegraphics[width=\linewidth]{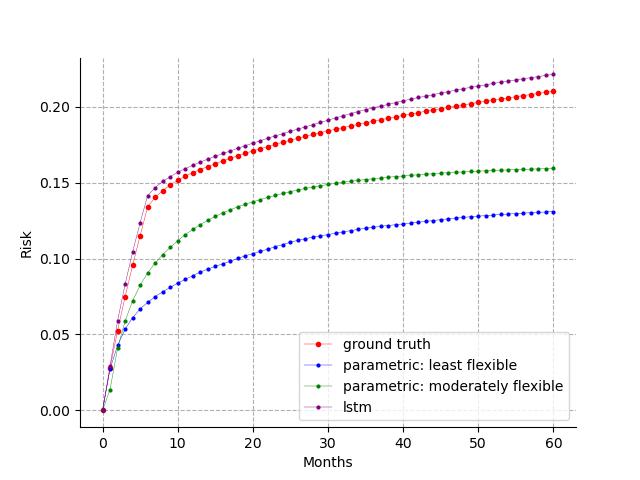} 
    \caption{Sample size = 1,000}
    \label{fig:sfig3}
\end{subfigure}
\hfill
\begin{subfigure}{0.49\textwidth}
    \includegraphics[width=\linewidth]{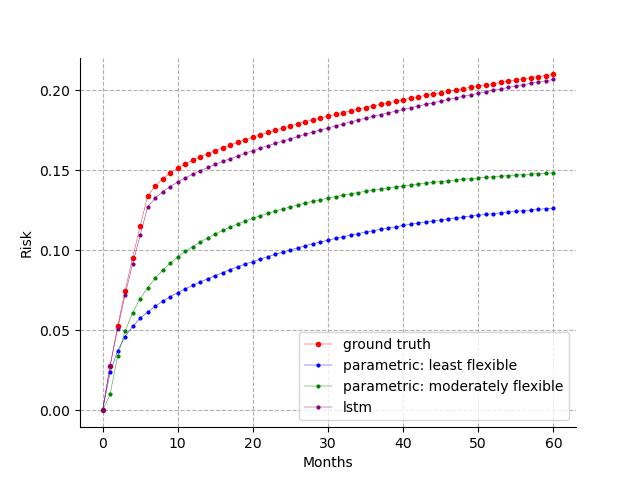} 
    \caption{Sample size = 10,000}
    \label{fig:sfig4}
\end{subfigure}
\caption{Ground truth and estimated risk under the 'natural course' using the parametric and DL-NICE methods on 'complex' time-dependency data}
\end{figure}

\begin{table}[H]
\centering
\resizebox{\textwidth}{!}{%
\begin{tabular}{|l|c|c|c|c|c|}
\hline
\textbf{Method} & \multicolumn{3}{c|}{\textbf{Bias in risk estimates} (\(\downarrow\) is better)} & \multicolumn{2}{c|}{\textbf{Bias in causal effect estimates} (\(\downarrow\) is better)} \\
\hline
 & \textbf{Natural course} & \textbf{Always treat} & \textbf{Never treat} & \textbf{Risk difference} & \textbf{Risk ratio} \\
\hline
Parametric 1* & 0.066 & 0.079 & 0.053 & 0.026 & 0.175 \\
Parametric 2** & 0.037 & 0.093 & 0.036 & 0.131 & 0.661 \\
DL-NICE (Ours) & \textbf{0.008} & \textbf{0.005} & \textbf{0.014} & \textbf{0.019} & \textbf{0.112} \\
\hline
\end{tabular}%
}
\vspace{0.5em} 
\caption{Bias in risk and causal effect estimates (comparing always vs. never treat; averaged over all 60 time points) using the parametric and DL-NICE methods on 'complex' time-dependency data with 1,000 individuals.
\footnotesize{* Parametric 1: Least flexible, ** Parametric 2: Moderately flexible.}}
\end{table}

\vspace{-2em} 

\begin{table}[H]
\centering
\resizebox{\textwidth}{!}{%
\begin{tabular}{|l|c|c|c|c|c|}
\hline
\textbf{Method} & \multicolumn{3}{c|}{\textbf{Bias in risk estimates} (\(\downarrow\) is better)} & \multicolumn{2}{c|}{\textbf{Bias in causal effect estimates} (\(\downarrow\) is better)} \\
\hline
 & \textbf{Natural course} & \textbf{Always treat} & \textbf{Never treat} & \textbf{Risk difference} & \textbf{Risk ratio} \\
\hline
Parametric 1* & 0.074 & 0.081 & 0.065 & 0.016 & \textbf{0.079} \\
Parametric 2** & 0.051 & 0.102 & 0.023 & 0.124 & 0.672 \\
DL-NICE (Ours) & \textbf{0.006} & \textbf{0.012} & \textbf{0.003} & \textbf{0.014} & 0.089 \\
\hline
\end{tabular}%
}
\vspace{0.5em} 
\caption{Bias in risk and causal effect estimates (comparing always vs. never treat; averaged over all 60 time points) using the parametric and DL-NICE methods on 'complex' time-dependency data with 10,000 individuals.
\footnotesize{* Parametric 1: Least flexible, ** Parametric 2: Moderately flexible.}}
\end{table}

\section{Conclusion}

We introduced a deep learning NICE g-formula estimator for causal inference from complex observational data that uses LSTMs to estimate the joint distribution of covariates and outcomes over time. This approach eliminates the need to fit separate parametric models for the conditional distribution of each covariate and specification of the functional form of covariate relationships. Our results showed that the DL-NICE estimator generally achieved lower bias in risk and causal effect estimates compared to parametric NICE, particularly in settings with complex temporal dependencies. Even in simpler settings, the DL-NICE estimator generally showed less bias.

However, the reduction in bias under the DL-NICE estimator was not uniform across scenarios, with the bias in the effect estimates being marginally higher in the 'simple' data with smaller sample size when using DL-NICE compared to parametric NICE. This finding suggests that the relative performance of the DL-NICE estimator may be sensitive to sample size and the complexity of the underlying data structure. Future work should explore statistical inference under the DL-NICE method, the conditions under which it may provide the most benefit over parametric NICE and model selection strategies.

\appendix
\section{Supplementary Material}

\setcounter{figure}{0} 
\setcounter{table}{0}
\setcounter{equation}{0}

\renewcommand{\thefigure}{S\arabic{figure}}
\renewcommand{\thetable}{S\arabic{table}}
\renewcommand{\theequation}{S\arabic{equation}}

\subsection{Data simulation design}
\label{sec:sim_design}

In our motivating example, baseline ($k=0$) was defined as meeting the following eligibility criteria for the first time: at least 18 years old, diagnosis of HIV-1 infection, starting ART for the first time, an HIV-RNA measurement while treatment-naïve that had to be detectable (>50 copies/ml) and no history of a cardiovascular event (myocardial infarction, stroke, or invasive cardiovascular procedure) or cancer. At baseline, $L_0$ included sex (binary, 1=male, 0=female), age (continuous) and smoking status (categorical, 0=never smoked, 1=currently smoking, 2=ex-smoker). The time-varying confounders in $L_k$ included a person’s HIV-RNA viral load, CD4 cell count, and an indicator of being overweight or obese at time $k$ based on a person’s body mass index (BMI >25). The binary treatment $A_k$ is an indicator for whether a person is using an ART regimen with an INSTI ($A_k=1$) or using an ART regimen without an INSTI ($A_k=0$). The outcome $Y_k$ is an indicator of whether the person experienced a cardiovascular event ($Y_k=1$). For each individual, data were simulated for up to 60 observation months. If and when an individual experienced an event ($Y_{k+1}=1$), follow up ends and there will be no subsequent observation months for that individual. The data generating processes in both the 'simple' and 'complex' time-dependency datasets are detailed below.\\

\subsubsection{Data generating processes for the "simple time dependency" dataset}
We first drew an unmeasured variable $U$ and simulated the time-fixed covariates sex ($L_{10}$), age ($L_{20}$) and smoking ($L_{30}$). We then generated the time-varying covariates in $L_k$ and $A_k$ and the survival outcome $Y_k$ at each time point. To generate increasing and decreasing trends in the time-varying covariates CD4 count and HIV-RNA, respectively, that would be observed over the first six months after ART is initiated, we used different data generating equations for the first six observation months and for the months after for these variables. For all time-varying covariates we also generated values at time $k=-1$ (pre-baseline), so that these could be used to generate the covariate values at time $k=0$ (baseline).  The data generating functions for each variable are shown below.

\subsection*{Baseline Covariates}
\begin{itemize}
    \item $Age \sim \text{TruncatedNormal}(\mu=50, \sigma=12, a=18, b=80)$
    \item $Sex \sim \text{Ber}(p=0.8)$
    \item $Smoking \sim \text{Multinomial}(1, [0.45, 0.4, 0.15])$
    \item $U \sim U(0.0, 1.0)$
\end{itemize}

\subsection*{Time-Varying Covariates}
\subsubsection*{CD4 Count:}
Pre-baseline value at time $k=-1$:
\begin{equation}
    \begin{aligned}
        CD4_{k=-1} & \sim \text{TruncatedNormal}(\mu=450, \sigma=100, a=350, b=800)
    \end{aligned}
\end{equation}
Time $0 \leq k \leq 5$:
\begin{equation}
    \begin{aligned}
        CD4_{k} & \sim \text{TruncatedNormal}(\mu, \sigma=100, a=350, b=800) \\
        \mu & \sim -0.8 \cdot U + 0.7 \cdot Sex - 0.8 \cdot Age - 0.05 \cdot Age^2 + 0.6 \cdot Smoking \\
            &+ 2 \cdot CD4_{k-1} + 0.005 \cdot CD4_{k-1}^2 - 1 \cdot RNA_{k-1} - 0.1 \cdot HighBMI_{k-1} \\
            &+ 0.05 \cdot Insti_{k-1} + 0.1 \cdot HighBMI_{k-1} \cdot RNA_{k-1} \\
            &+ 0.08 \cdot HighBMI_{k-1} \cdot Sex - 0.1 \cdot CD4_{k-1} \cdot Age
    \end{aligned}
\end{equation}\\
At time $k=0$, the coefficient for $Insti_{k-1}$ was set to 0 as there is no treatment before baseline. \\
\\
From time $k \geq 6$:
\begin{equation}
    \begin{aligned}
        CD4_{k} & \sim \text{TruncatedNormal}(\mu, \sigma=80, a=400, b=800) \\
        \mu & \sim -0.8 \cdot U + 0.7 \cdot Sex - 0.8 \cdot Age - 0.05 \cdot Age^2 + 0.6 \cdot Smoking \\
            &+ 1.8 \cdot CD4_{k-1} + 0.008 \cdot CD4_{k-1}^2 - 1 \cdot RNA_{k-1} - 0.1 \cdot HighBMI_{k-1} \\
            &+ 0.05 \cdot Insti_{k-1} + 0.1 \cdot HighBMI_{k-1} \cdot RNA_{k-1} \\
            &+ 0.08 \cdot HighBMI_{k-1} \cdot Sex - 0.1 \cdot CD4_{k-1} \cdot Age
    \end{aligned}
\end{equation}

\subsubsection*{HIV RNA:}
Pre-baseline value at time $k=-1$:
\begin{equation}
    \begin{aligned}
        RNA_{k=-1} & \sim \text{TruncatedNormal}(\mu=60, \sigma=30, a=40, b=90)
    \end{aligned}
\end{equation}
Time $0 \leq k \leq 5$:
\begin{equation}
    \begin{aligned}
        RNA_{k} & \sim \text{TruncatedNormal}(\mu, \sigma=30, a=40, b=80) \\
        \mu & \sim 0.5 \cdot U + 0.5 \cdot Sex + 0.3 \cdot Age + 0.006 \cdot Age^2 + 0.8 \cdot Smoking \\
            &- 0.5 \cdot CD4_{k-1} - 0.0001 \cdot CD4_{k-1}^2 + 2 \cdot RNA_{k-1} + 0.3 \cdot HighBMI_{k-1} \\
            &+ 0.05 \cdot INSTI_{k-1} + 0.08 \cdot HighBMI_{k-1} \cdot RNA_{k-1} \\
            &+ 0.1 \cdot HighBMI_{k-1} \cdot Sex - 0.01 \cdot CD4_{k-1} \cdot Age
    \end{aligned}
\end{equation}
From time $k \geq 6$:
\begin{equation}
    \begin{aligned}
        RNA_{k} & \sim \text{TruncatedNormal}(\mu, \sigma=20, a=20, b=70) \\
        \mu & \sim 0.5 \cdot U + 0.5 \cdot Sex + 0.3 \cdot Age + 0.006 \cdot Age^2 + 0.8 \cdot Smoking \\
            &- 0.5 \cdot CD4_{k-1} - 0.0001 \cdot CD4_{k-1}^2 + 1.8 \cdot RNA_{k-1} + 0.3 \cdot HighBMI_{k-1} \\
            &+ 0.05 \cdot INSTI_{k-1} + 0.08 \cdot HighBMI_{k-1} \cdot RNA_{k-1} \\
            &+ 0.1 \cdot HighBMI_{k-1} \cdot Sex - 0.01 \cdot CD4_{k-1} \cdot Age
    \end{aligned}
\end{equation}

\subsubsection*{BMI $>25$:}
\begin{equation}
    \begin{aligned}
        HighBMI_{k} & \sim \text{Ber}\left(\text{expit}(\mu)\right) \\
        \mu & \sim -8 -2 \cdot U + 0.03 \cdot Sex + 0.01 \cdot Age + 0.0001 \cdot Age^2 + 0.04 \cdot Smoking \\
            &- 0.0001 \cdot CD4_{k-1} + 0.001 \cdot RNA_{k-1} + 10 \cdot HighBMI_{k-1} \\
            &+ 5 \cdot INSTI_{k-1} + 0.001 \cdot HighBMI_{k-1} \cdot RNA_{k-1} \\
            &+ 0.004 \cdot HighBMI_{k-1} \cdot Sex + 0.00001 \cdot CD4_{k-1} \cdot Age
    \end{aligned}
\end{equation}

\subsection*{Treatment}
\begin{equation}
    \begin{aligned}
        INSTI_{k} & \sim \text{Ber}\left(\text{expit}(\mu)\right) \\
        \mu & \sim -4.5 + 0.5 \cdot Sex + 0.01 \cdot Age + 0.0001 \cdot Age^2 + 0.1 \cdot Smoking \\
            &+ 0.001 \cdot CD4_{k} + 0.01 \cdot RNA_{k} - 7 \cdot HighBMI_{k} \\
            &+ 10 \cdot INSTI_{k-1} + 0.0001 \cdot HighBMI_{k} \cdot RNA_{k} \\
            &+ 0.001 \cdot HighBMI_{k} \cdot Sex + 0.00001 \cdot CD4_{k} \cdot Age
    \end{aligned}
\end{equation}

\subsection*{Outcome}
\begin{equation}
    \begin{aligned}
        Y_k & \sim \text{Ber}\left(\text{expit}(\mu)\right) \\
        \mu & = -0.08 \cdot U + 0.005 \cdot Sex + 0.015 \cdot Age + 0.00000005 \cdot Age^2 + 0.025 \cdot Smoking \\
            & - 0.015 \cdot CD4_{k} + 0.03 \cdot RNA_{k} + 0.0000004 \cdot RNA_{k}^2 + 0.1 \cdot HighBMI_{k} \\
            & + 0.09 \cdot INSTI_{k}
    \end{aligned}
\end{equation}

\subsubsection{Data generating processes for the "complex time dependency" dataset} 
The data generation functions for the unmeasured confounder $U$ and the baseline covariates in $L_0$ were the same as in the simple time dependency scenario. The time-varying covariates in $L_k$, treatment $A_k$ and the outcome $Y_k$ were generated to depend on a function of the entire history of covariates and treatment, defined as the cumulative average of values over the previous 6, 7-24 and over >24 time points from the current time point $k$ for $L_k$ and the cumulative average over all previous time points after baseline for $A_k$. As in the 'simple' time-dependency dataset, we used different data generating equations for the first six observation months and for the months after for the time-varying covariates CD4 count and HIV-RNA. For all time-varying covariates we also generated values for 30 pre-baseline observation months  ($k=-30 to k=-1$) to be used to generate the covariate values from time $k=0$ (baseline) onwards.  The data generating functions for each covariate $L_k$, $A_k$ and the outcome $Y_k$ are shown below.

\subsection*{Time-Varying Covariates}
\subsubsection*{CD4 Count:}
Initial value at time $k=-30$:
\begin{equation}
    \begin{aligned}
        CD4_{k=-30} & \sim \text{TruncatedNormal}(\mu=450, \sigma=100, a=350, b=800)
    \end{aligned}
\end{equation}
Value evolution from $k=-30$ to $k=-1$:
\begin{equation}
    \begin{aligned}
        CD4_{k} & = CD4_{k-1} \cdot 0.995 \quad \text{for} \quad k=-29 \quad \text{to} \quad k=-1
    \end{aligned}
\end{equation}
Time $0 \leq k \leq 5$:
\begin{equation}
    \begin{aligned}
        CD4_{k} & \sim \text{TruncatedNormal}(\mu, \sigma=100, a=350, b=800) \\
        \mu & = -2 \cdot U + 0.1 \cdot Sex - 1 \cdot Age - 0.05 \cdot Age^2 + 0.3 \cdot Smoking \\
            & + 1 \cdot CD4_{k-1} + 0.005 \cdot CD4_{k-1}^2 + 0.8 \cdot \text{mean}(CD4_{k-6:k-1}) \\
            & + 0.6 \cdot \text{mean}(CD4_{k-24:k-7}) + 0.5 \cdot \text{mean}(CD4_{k-25:k=-30}) \\
            & - 0.8 \cdot RNA_{k-1} - 0.5 \cdot \text{mean}(RNA_{k-6:k-1}) - 0.3 \cdot \text{mean}(RNA_{k-24:k-7})\\
            & - 0.1 \cdot \text{mean}(RNA_{k-25:k=-30}) - 0.1 \cdot HighBMI_{k-1} - 0.08 \cdot \text{mean}(HighBMI_{k-6:k-1}) \\
            & - 0.04 \cdot \text{mean}(HighBMI_{k-24:k-7}) - 0.02 \cdot \text{mean}(HighBMI_{k-25:k=-30}) \\
            & + 0.05 \cdot INSTI_{k-1} + 0.02 \cdot \text{mean}(INSTI_{k-1:k=0}) + 0.1 \cdot HighBMI_{k-1} \cdot RNA_{k-1} \\
            & + 0.08 \cdot HighBMI_{k-1} \cdot Sex - 0.1 \cdot CD4_{k-1} \cdot Age
    \end{aligned}
\end{equation}
From time $k \geq 6$:
\begin{equation}
    \begin{aligned}
        CD4_{k} & \sim \text{TruncatedNormal}(\mu, \sigma=80, a=400, b=800) \\
        \mu & = -2 \cdot U + 0.1 \cdot Sex - 1 \cdot Age - 0.05 \cdot Age^2 + 0.3 \cdot Smoking \\
            & + 1.8 \cdot CD4_{k-1} + 0.008 \cdot CD4_{k-1}^2 + 1 \cdot \text{mean}(CD4_{k-6:k-1}) \\
            & + 0.5 \cdot \text{mean}(CD4_{k-24:k-7}) + 0.2 \cdot \text{mean}(CD4_{k-25:k=-30}) \\
            & - 0.8 \cdot RNA_{k-1} - 0.5 \cdot \text{mean}(RNA_{k-6:k-1}) - 0.3 \cdot \text{mean}(RNA_{k-24:k-7})\\
            & - 0.1 \cdot \text{mean}(RNA_{k-25:k=-30}) - 0.1 \cdot HighBMI_{k-1} - 0.08 \cdot \text{mean}(HighBMI_{k-6:k-1}) \\
            & - 0.04 \cdot \text{mean}(HighBMI_{k-24:k-7}) - 0.02 \cdot \text{mean}(HighBMI_{k-25:k=-30}) \\
            & + 0.05 \cdot INSTI_{k-1} + 0.02 \cdot \text{mean}(INSTI_{k-1:k=0}) + 0.1 \cdot HighBMI_{k-1} \cdot RNA_{k-1} \\
            & + 0.08 \cdot HighBMI_{k-1} \cdot Sex - 0.1 \cdot CD4_{k-1} \cdot Age
    \end{aligned}
\end{equation}

\subsubsection*{HIV RNA:}
Initial value at time $k=-30$:
\begin{equation}
    \begin{aligned}
        RNA_{k=-30} & \sim \text{TruncatedNormal}(\mu=60, \sigma=30, a=40, b=90)
    \end{aligned}
\end{equation}
Value evolution from $k=-30$ to $k=-1$:
\begin{equation}
    \begin{aligned}
        RNA_{k} & = RNA_{k-1} \cdot 1.01 \quad \text{for} \quad k=-29 \quad \text{to} \quad k=-1
    \end{aligned}
\end{equation}
Time $0 \leq k \leq 5$:
\begin{equation}
    \begin{aligned}
        RNA_{k} & \sim \text{TruncatedNormal}(\mu, \sigma=30, a=40, b=80) \\
        \mu & = 2 \cdot U + 1 \cdot Sex + 1.5 \cdot Age + 0.006 \cdot Age^2 + 0.5 \cdot Smoking \\
            & -0.5 \cdot CD4_{k-1} - 0.0001 \cdot CD4_{k-1}^2 - 0.2 \cdot \text{mean}(CD4_{k-6:k-1}) \\
            & - 0.05 \cdot \text{mean}(CD4_{k-24:k-7}) - 0.01 \cdot \text{mean}(CD4_{k-25:k=-30}) \\
            & + 6 \cdot RNA_{k-1} + 5 \cdot \text{mean}(RNA_{k-6:k-1}) + 3 \cdot \text{mean}(RNA_{k-24:k-7})\\
            & + 2 \cdot \text{mean}(RNA_{k-25:k=-30}) + 0.3 \cdot HighBMI_{k-1} + 0.1 \cdot \text{mean}(HighBMI_{k-6:k-1}) \\
            & + 0.06 \cdot \text{mean}(HighBMI_{k-24:k-7}) + 0.03 \cdot \text{mean}(HighBMI_{k-25:k=-30}) \\
            & + 0.05 \cdot INSTI_{k-1} + 0.01 \cdot \text{mean}(INSTI_{k-1:k=0}) + 0.08 \cdot HighBMI_{k-1} \cdot RNA_{k-1} \\
            & + 0.1 \cdot HighBMI_{k-1} \cdot Sex - 0.01 \cdot CD4_{k-1} \cdot Age
    \end{aligned}
\end{equation}
From time $k \geq 6$:
\begin{equation}
    \begin{aligned}
         RNA_{k} & \sim \text{TruncatedNormal}(\mu, \sigma=20, a=20, b=70) \\
        \mu & = 2 \cdot U + 1 \cdot Sex + 1.5 \cdot Age + 0.006 \cdot Age^2 + 0.5 \cdot Smoking \\
            & -0.5 \cdot CD4_{k-1} - 0.0001 \cdot CD4_{k-1}^2 - 0.2 \cdot \text{mean}(CD4_{k-6:k-1}) \\
            & - 0.05 \cdot \text{mean}(CD4_{k-24:k-7}) - 0.01 \cdot \text{mean}(CD4_{k-25:k=-30}) \\
            & + 6 \cdot RNA_{k-1} + 5 \cdot \text{mean}(RNA_{k-6:k-1}) + 3 \cdot \text{mean}(RNA_{k-24:k-7})\\
            & + 2 \cdot \text{mean}(RNA_{k-25:k=-30}) + 0.3 \cdot HighBMI_{k-1} + 0.1 \cdot \text{mean}(HighBMI_{k-6:k-1}) \\
            & + 0.06 \cdot \text{mean}(HighBMI_{k-24:k-7}) + 0.03 \cdot \text{mean}(HighBMI_{k-25:k=-30}) \\
            & + 0.05 \cdot INSTI_{k-1} + 0.01 \cdot \text{mean}(INSTI_{k-1:k=0}) + 0.08 \cdot HighBMI_{k-1} \cdot RNA_{k-1} \\
            & + 0.1 \cdot HighBMI_{k-1} \cdot Sex - 0.01 \cdot CD4_{k-1} \cdot Age
    \end{aligned}
\end{equation}

\subsubsection*{BMI $>25$:}
\begin{equation}
    \begin{aligned}
        HighBMI_{k} & \sim \text{Ber}\left(\text{expit}(\mu)\right) \\
        \mu & = -6.5 -1 \cdot U - 0.6 \cdot Sex + 0.01 \cdot Age + 0.0001 \cdot Age^2 - 0.04 \cdot Smoking \\
            & - 0.0001 \cdot CD4_{k-1} + 0.000001 \cdot CD4_{k-1}^2 - 0.001 \cdot \text{mean}(CD4_{k-6:k-1}) \\
            & - 0.0001 \cdot \text{mean}(CD4_{k-24:k-7}) - 0.001 \cdot \text{mean}(CD4_{k-25:k=-30}) \\
            & + 0.01 \cdot RNA_{k-1} + 0.01 \cdot \text{mean}(RNA_{k-6:k-1}) + 0.007 \cdot \text{mean}(RNA_{k-24:k-7})\\
            & + 0.006 \cdot \text{mean}(RNA_{k-25:k=-30}) + 4.5 \cdot HighBMI_{k-1} + 3 \cdot \text{mean}(HighBMI_{k-6:k-1}) \\
            & + 1.6 \cdot \text{mean}(HighBMI_{k-24:k-7}) + 1 \cdot \text{mean}(HighBMI_{k-25:k=-30}) \\
            & + 2 \cdot INSTI_{k-1} + 1 \cdot \text{mean}(INSTI_{k-1:k=0}) 
    \end{aligned}
\end{equation}

\subsection*{Treatment}
\begin{equation}
    \begin{aligned}
        INSTI_{k} & \sim \text{Ber}\left(\text{expit}(\mu)\right) \\
        \mu & = -4 + 0.5 \cdot Sex + 0.05 \cdot Age + 0.00005 \cdot Age^2 + 0.2 \cdot Smoking \\
            & - 0.001 \cdot CD4_{k} + 0.0000001 \cdot CD4_{k}^2 - 0.0001 \cdot \text{mean}(CD4_{k-6:k-1}) \\
            & + 0.001 \cdot RNA_{k} + 0.0003 \cdot \text{mean}(RNA_{k-6:k-1}) \\
            & - 3 \cdot HighBMI_{k} - 2 \cdot \text{mean}(HighBMI_{k-6:k-1}) \\
            & - 1.3 \cdot \text{mean}(HighBMI_{k-24:k-7}) - 0.8 \cdot \text{mean}(HighBMI_{k-25:k=-30}) \\
            & + 6 \cdot INSTI_{k-1} + 4 \cdot \text{mean}(INSTI_{k-1:k=0}) 
    \end{aligned}
\end{equation}

\subsection*{Outcome}
\begin{equation}
    \begin{aligned}
        Y_{k} & \sim \text{Ber}\left(\text{expit}(\mu)\right) \\
        \mu & = -0.05 \cdot U + 0.007 \cdot Sex + 0.02 \cdot Age + 0.00000005 \cdot Age^2 + 0.03 \cdot Smoking \\
            & - 0.009 \cdot CD4_{k} - 0.008 \cdot \text{mean}(CD4_{k-6:k-1}) - 0.006 \cdot \text{mean}(CD4_{k-24:k-7}) \\
            & - 0.004 \cdot \text{mean}(CD4_{k-25:k=-30}) + 0.045 \cdot RNA_{k} + 0.0000004 \cdot RNA_{k}^2 \\
            & + 0.03 \cdot \text{mean}(RNA_{k-6:k-1}) + 0.025 \cdot \text{mean}(RNA_{k-24:k-7}) + 0.02 \cdot \text{mean}(RNA_{k-25:k=-30}) \\
            & + 0.14 \cdot HighBMI_{k} + 0.11 \cdot \text{mean}(HighBMI_{k-6:k-1}) + 0.08 \cdot \text{mean}(HighBMI_{k-24:k-7}) \\
            & + 0.06 \cdot \text{mean}(HighBMI_{k-25:k=-30}) + 0.13 \cdot INSTI_{k} + 0.11 \cdot \text{mean}(INSTI_{k-1:k=0})
    \end{aligned}
\end{equation}

\subsection{Identifying assumptions and the g-formula}
\label{sec:ident_assump}

Let $Y_k^g$ and $L_k^g$ denote the counterfactual outcome and vector of confounders, respectively, at time $k$ had an individual followed a deterministic strategy $g$ $(\forall{k} = 1,\dots,K)$. For simplicity, we will again ignore censoring. Consider the following assumptions for a particular strategy $g$:\\
\emph{1. Exchangeability}:
\begin{equation*} 
(Y_{k+1}^g,...,Y_k^g) \perp (A_k) | \bar{L}_k=\bar{l}_k,\bar{A}_{k-1}=\bar{a}_{k-1}^g,Y_k=0
\end{equation*}
\emph{2. Positivity}:
\begin{equation*} 
f_{\bar{L}_k,\bar{A}_{k-1},Y_k}(\bar{l}_k,\bar{a}_{k-1}^g,0,0) >0 \Rightarrow f_{{A}_k|\bar{L}_{k},\bar{A}_{k-1},Y_k}(a_k^g|\bar{l}_k,\bar{a}_{k-1}^g,0,0) >0
\end{equation*}
\emph{3. Consistency}: If $\bar{A}_k=\bar{A}_k^g$ then $\bar{Y}_{k+1}^g$ and $\bar{L}_k=\bar{L}_k^g$.\\

Let $G$ be the set of all deterministic interventions $g$. Provided that these assumptions hold for a subset of $g \in G $ that are observable under $f^{int}(a_k|Y_k=0,\bar{l}_{k},\bar{a}_{k-1}) (\forall{k=0,...,K-1})$ then the risk by $K$ had all subjects been assigned treatment according to $f^{int}(a_k|Y_k=0,\bar{l}_{k},\bar{a}_{k-1})$ can be estimated from the observational data and written as the g-formula, an expectation weighted by the joint density of covariates:
\begin{equation}\label{full_gform}
\begin{split}
\sum_{\forall{\bar{a}_{K-1}}} \sum_{\forall{\bar{l}_{K-1}}} \sum_{k=1}^{K} P(Y_k=1|Y_{k-1}=0,\bar{L}_{k-1} =\bar{l}_{k-1},\bar{A}_{k-1}=\bar{a}_{k-1}) \times \\
\prod_{s=0}^{k-1} P(Y_s=0||Y_{s-1}=0,\bar{L}_{s-1} =
\bar{l}_{s-1},\bar{A}_{s-1}=\bar{a}_{s-1}) \\
f(l_s|Y_s=0,\bar{l}_{s-1},\bar{a}_{s-1})f^{int}(a_s|Y_s=0,\bar{l}_s,\bar{a}_{s-1}) 
\end{split}
\end{equation}

where $P(Y_k=1|Y_{k-1}=0,\bar{L}_{k-1}=\bar{l}_{k-1},\bar{A}_{k-1}=\bar{a}_{k-1}$ and $f(l_k|Y_{k-1}=0,\bar{l}_{k-1},\bar{a}_{k-1})$ are the observed discrete-time hazards of the outcome and the joint density of the confounders at time $k$, respectively, conditional on past treatment, confounders and survival through time $k-1$ (as we are ignoring censoring in this study, we are not additionally conditioning on being uncensored through time $k$ but this is typically included in the g-formula for survival outcomes). For notational simplicity, the above expression assumes that all covariates are discrete; otherwise, the sum would be replaced by an integral.

\subsection{Treatment strategies and causal contrast}
\label{sec:treat_strat}
A treatment strategy is a rule that assigns treatment at each time $k$ as an independent draw from an intervention distribution $f^{int}(a_k|Y_k=0,\bar{l}_{k},\bar{a}_{k-1})$ that may, at most, depend on $(\bar{l}_k,\bar{a}_{k-1})$, which is a realization of $(\bar{L}_k,\bar{A}_{k-1})$. Treatment strategies can be either deterministic or random. A treatment strategy is deterministic if at each time point $k (k=0,\dots,K)$, $f^{int}(a_k|Y_k=0,\bar{l}_{k},\bar{a}_{k-1})$ either equals 0 or 1 for all $(\bar{l}_k,\bar{a}_{k-1})$. Otherwise, the treatment strategy is random. We denote $g$ as a deterministic treatment strategy, and $a_k^g = g_k(\bar{a}_{k-1}^g,\bar{l}_k)$ the value of treatment assigned at time k under $g (\forall{k=0,\dots,K-1})$. In the present study, we consider a deterministic static treatment strategy. A treatment strategy is static if the rule for assigning treatment at each time point does not depend on past covariates. The treatment strategy in our study is “initiate ART containing INSTI at time 0 and continue to treat with this combination during the study” (or “always treat”), which corresponds to $a_k^g=1$, for all $k$ and for any $(\bar{a}_{k-1},\bar{l}_k)$. We estimate the causal effect on the risk (cumulative incidence) by time $K$ of the “always treat” treatment strategy compared to a “never treat” strategy ($a_k^g=0$, for all $k$ and for any $(\bar{a}_{k-1},\bar{l}_k)$), which in our motivating example is defined as “initiate ART \emph{not} containing INSTI at time 0 and continue with this strategy during the study”. That is, we estimate a contrast in the risk had all individuals been assigned treatment (i.e. initiated ART containing INSTI) vs. had no one been assigned this treatment strategy (had no one taken ART containing INSTI).

\clearpage 
\begin{landscape}
\centering
\subsection{Hyperparameter search and settings}
\label{sec:hyper_param}

\begin{table}[H]
\centering
\resizebox{1.4\textwidth}{!}{%
\begin{tabular}{|l|c|c|c|c|c|c|c|c|c|}
\hline
\textbf{Hyperparameter} & \makecell{\textbf{Feature}\\\textbf{dimension}} & \makecell{\textbf{Hidden size}} & \makecell{\textbf{Layers}\\\textbf{num}} & \makecell{\textbf{Dropout}\\\textbf{rate}} & \makecell{\textbf{Learning}\\\textbf{rate}} & \makecell{\textbf{Batch}\\\textbf{size}} & \makecell{\textbf{Weight}\\\textbf{decay}} & \makecell{\textbf{Max}\\\textbf{epochs}} & \makecell{\textbf{Patience}\\\textbf{epoch}} \\
\hline
\textbf{Covariate LSTM} & \{128, 512\} & \{64, 128, 256\} & \{2, 3, 4\} & Uniform(0, 0.5) & log-uniform(1e-5, 1e-3) & \{512, 1024\} & log-uniform(1e-6, 1e-3) & 5000 & 50 \\
\textbf{Outcome LSTM}   & \{128, 512\} & \{64, 128, 256\} & \{2, 3, 4\} & Uniform(0, 0.5) & log-uniform(1e-5, 1e-3) & \{512, 1024\} & log-uniform(1e-6, 1e-3) & 1000 & 50 \\
\hline
\end{tabular}%
}
\vspace{0.5em}
\caption{Hyperparameter search space for the LSTMs in the DL-NICE estimator.}
\end{table}

\begin{table}[H]
\centering
\begin{tabular}{|l|c|c|c|c|c|c|c|c|c|}
\hline
\textbf{Model type} & \makecell{\textbf{Hidden size}\\\textbf{of LSTM}} & \makecell{\textbf{Layers}\\\textbf{num}} & \makecell{\textbf{Feature}\\\textbf{dimension}} & \textbf{Dropout} & \makecell{\textbf{Epoch}\\\textbf{(best)}} & \makecell{\textbf{Patience}\\\textbf{parameter}} & \makecell{\textbf{Batch}\\\textbf{size}} & \makecell{\textbf{Learning}\\\textbf{rate}} & \makecell{\textbf{Weight}\\\textbf{decay}} \\
\hline
\multicolumn{10}{|c|}{\textbf{Sample size = 10,000}} \\
\hline
Covariate LSTM & 64 & 3 & 128 & 0.19 & 5000 (588) & 50 & 1024 & 4.1e-4 & 3.3e-6 \\
Outcome LSTM   & 64 & 4 & 128 & 0.15 & 1000 (1000) & 50 & 512  & 4.7e-5 & 7.4e-4 \\
\hline
\multicolumn{10}{|c|}{\textbf{Sample size = 1,000}} \\
\hline
Covariate LSTM & 64 & 2 & 512 & 0.45 & 5000 (716) & 50 & 512  & 3.8e-4 & 4.1e-5 \\
Outcome LSTM   & 128 & 2 & 128 & 0.23 & 1000 (455) & 50 & 512  & 6.7e-4 & 3.6e-5 \\
\hline
\end{tabular}
\vspace{0.5em} 
\caption{Hyperparameter settings for LSTMs for 'simple' time-dependency dataset with sample sizes 10,000 and 1,000.}
\end{table}

\begin{table}[H]
\centering
\begin{tabular}{|l|c|c|c|c|c|c|c|c|c|}
\hline
\textbf{Model type} & \makecell{\textbf{Hidden size}\\\textbf{of LSTM}} & \makecell{\textbf{Layers}\\\textbf{num}} & \makecell{\textbf{Feature}\\\textbf{dimension}} & \textbf{Dropout} & \makecell{\textbf{Epoch}\\\textbf{(best)}} & \makecell{\textbf{Patience}\\\textbf{parameter}} & \makecell{\textbf{Batch}\\\textbf{size}} & \makecell{\textbf{Learning}\\\textbf{rate}} & \makecell{\textbf{Weight}\\\textbf{decay}} \\
\hline
\multicolumn{10}{|c|}{\textbf{Sample size = 10,000}} \\
\hline
Covariate LSTM & 64 & 2 & 128 & 0.36 & 5000 (558) & 50 & 512  & 4.4e-4 & 4.7e-5 \\
Outcome LSTM   & 256 & 2 & 512 & 0.30 & 1000 (712) & 50 & 1024 & 2.4e-5 & 4.4e-4 \\
\hline
\multicolumn{10}{|c|}{\textbf{Sample size = 1,000}} \\
\hline
Covariate LSTM & 64 & 2 & 512 & 0.36 & 5000 (2239) & 50 & 1024 & 1.8e-4 & 3.1e-4 \\
Outcome LSTM   & 128 & 2 & 512 & 0.19 & 1000 (442) & 50 & 1024 & 5.5e-4 & 1.4e-6 \\
\hline
\end{tabular}
\vspace{0.5em} 
\caption{Hyperparameter settings for LSTMs for 'complex' time-dependency dataset with sample sizes 10,000 and 1,000.}
\end{table}

\end{landscape}

\subsection{Estimated risks and bias in risk and effect estimates over time }


\subsubsection{'Simple' time-dependency scenario}
\label{sec:risks_simple}

\begin{figure}[htbp]
\centering
\begin{subfigure}{0.49\textwidth}
    \includegraphics[width=\linewidth]{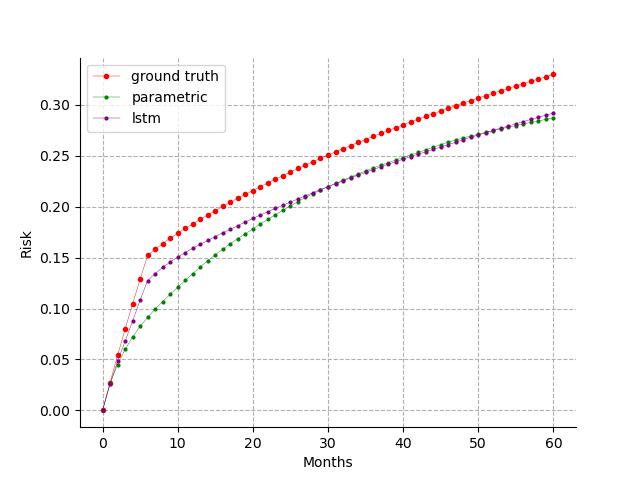} 
    \caption{Risk under always treat}
    \label{fig:sfig5}
\end{subfigure}
\hfill
\begin{subfigure}{0.49\textwidth}
    \includegraphics[width=\linewidth]{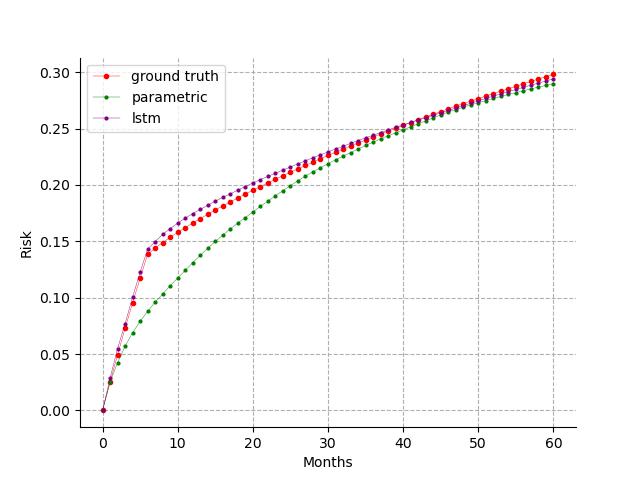} 
    \caption{Risk under never treat}
    \label{fig:sfig6}
\end{subfigure}

\caption{Ground truth and estimated risks under interventions in dataset with 1,000 individuals}
\end{figure}

\FloatBarrier 

\begin{figure}[htbp]
\centering
\begin{subfigure}{0.49\textwidth}
    \includegraphics[width=\linewidth]{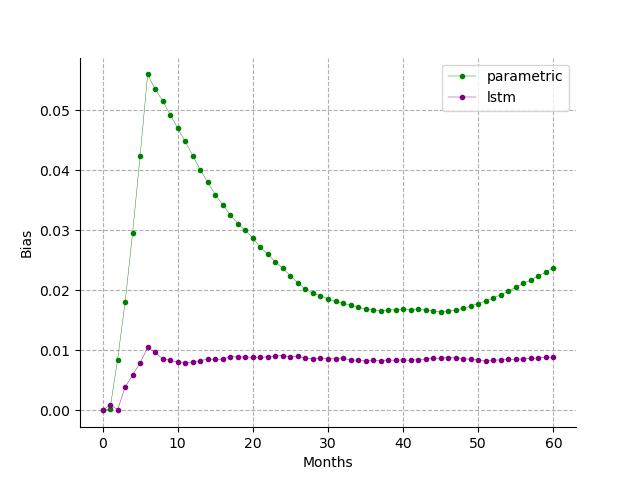} 
    \caption{Bias in risk under natural course}
    \label{fig:sfig7}
\end{subfigure}
\hfill
\begin{subfigure}{0.49\textwidth}
    \includegraphics[width=\linewidth]{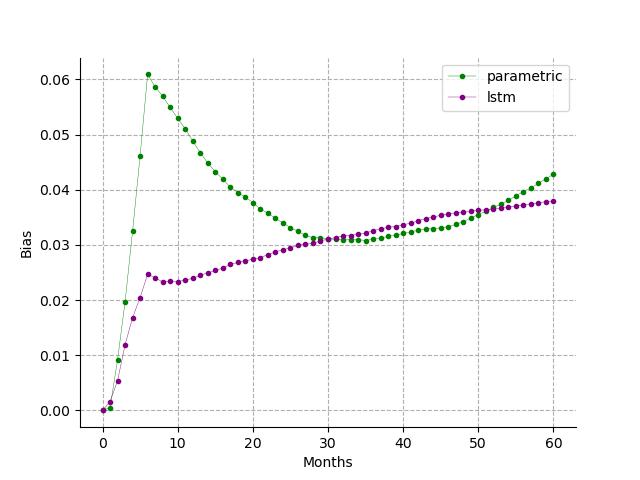} 
    \caption{Bias in risk under always treat}
    \label{fig:sfig8}
\end{subfigure}

\begin{subfigure}{0.49\textwidth} 
    \includegraphics[width=\linewidth]{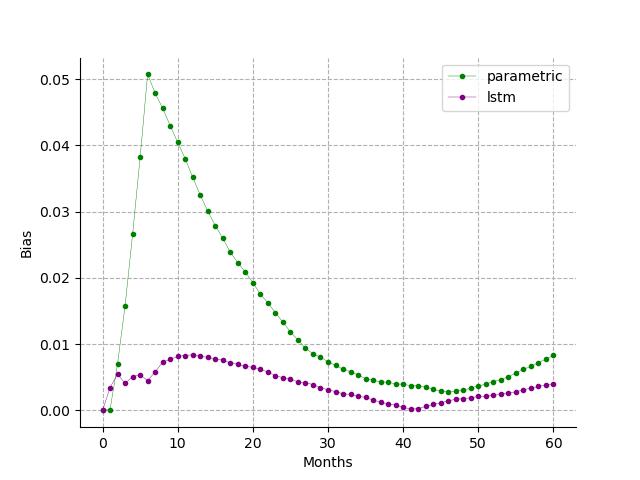} 
    \caption{Bias in risk under never treat}
    \label{fig:sfig9}
\end{subfigure}

\caption{Bias in risk in dataset with 1,000 individuals}
\end{figure}

\FloatBarrier 

\begin{figure}[htbp]
\centering
\begin{subfigure}{0.49\textwidth}
    \includegraphics[width=\linewidth]{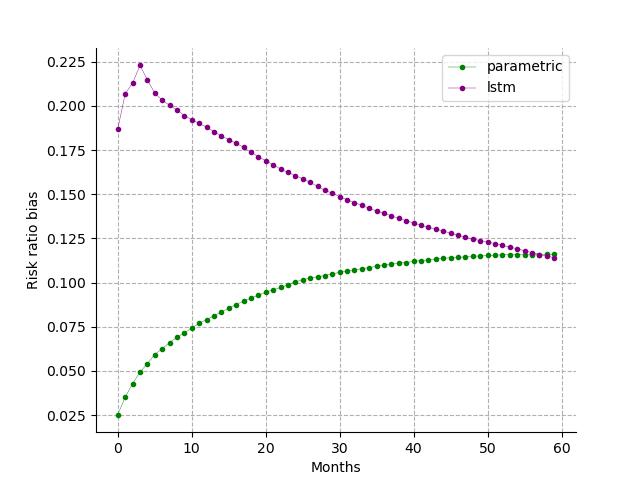} 
    \caption{Bias in risk ratio}
    \label{fig:sfig10}
\end{subfigure}
\hfill
\begin{subfigure}{0.49\textwidth}
    \includegraphics[width=\linewidth]{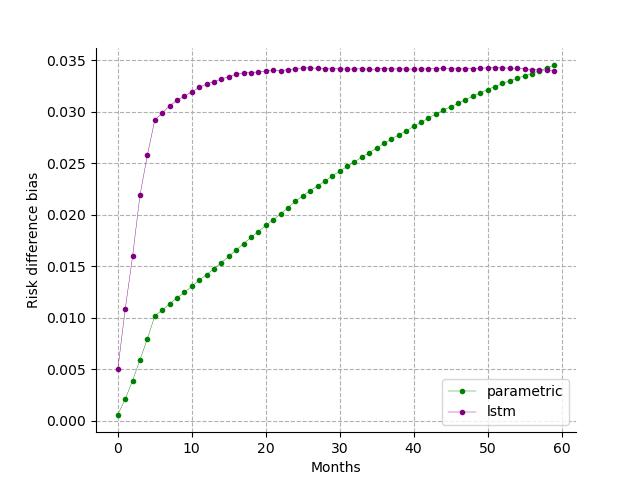} 
    \caption{Bias in risk difference}
    \label{fig:sfig11}
\end{subfigure}

\caption{Bias in effect estimates (always vs. never treat interventions) in 1,000 sample dataset}
\end{figure}

\FloatBarrier

\begin{figure}[p] 
\centering
\begin{subfigure}{0.49\textwidth}
    \includegraphics[width=\linewidth]{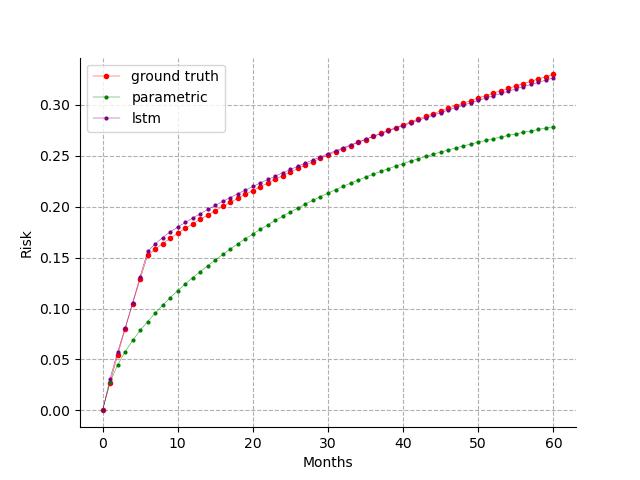} 
    \caption{Risk under always treat}
    \label{fig:sfig12}
\end{subfigure}
\hfill
\begin{subfigure}{0.49\textwidth}
    \includegraphics[width=\linewidth]{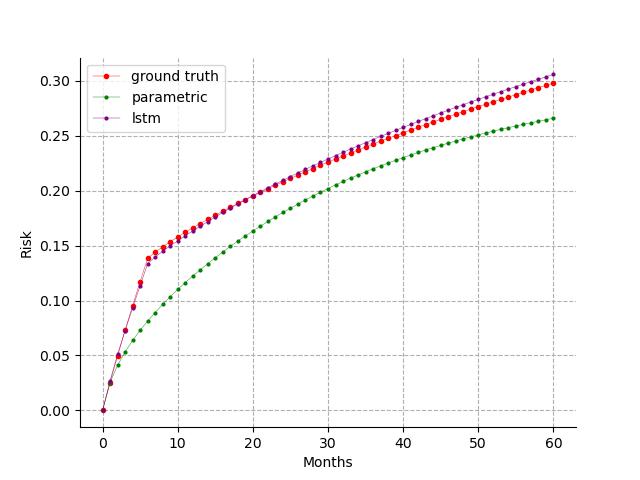} 
    \caption{Risk under never treat}
    \label{fig:sfig13}
\end{subfigure}
\caption{Ground truth and estimated risks under interventions in dataset with 10,000 individuals}
\end{figure}

\begin{figure}[p]
\centering
\begin{subfigure}{0.49\textwidth}
    \includegraphics[width=\linewidth]{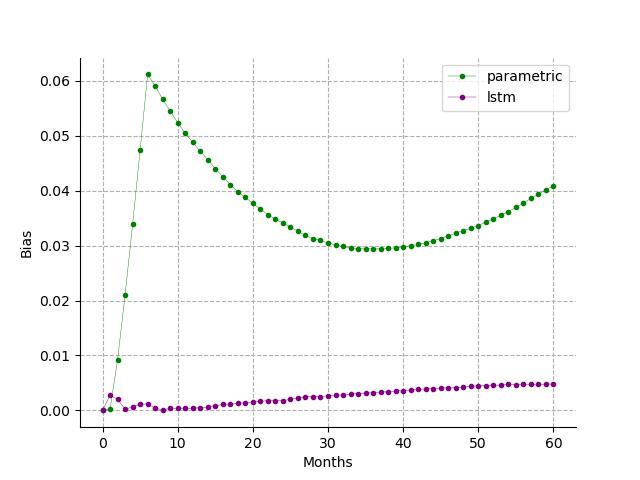} 
    \caption{Bias in risk under natural course}
    \label{fig:sfig14}
\end{subfigure}
\hfill
\begin{subfigure}{0.49\textwidth}
    \includegraphics[width=\linewidth]{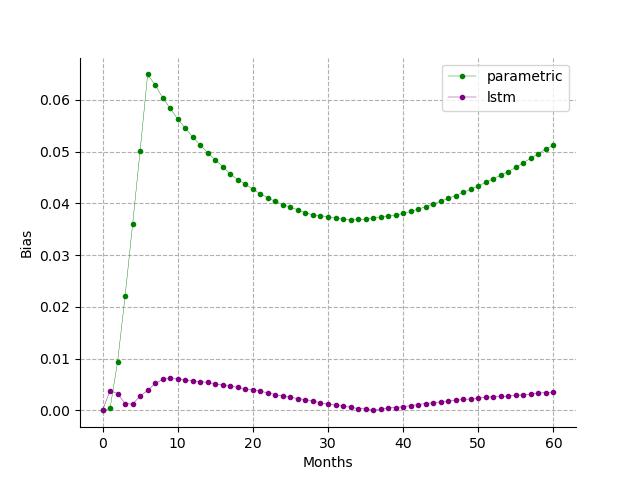} 
    \caption{Bias in risk under always treat}
    \label{fig:sfig15}
\end{subfigure}
\begin{subfigure}{0.49\textwidth} 
    \includegraphics[width=\linewidth]{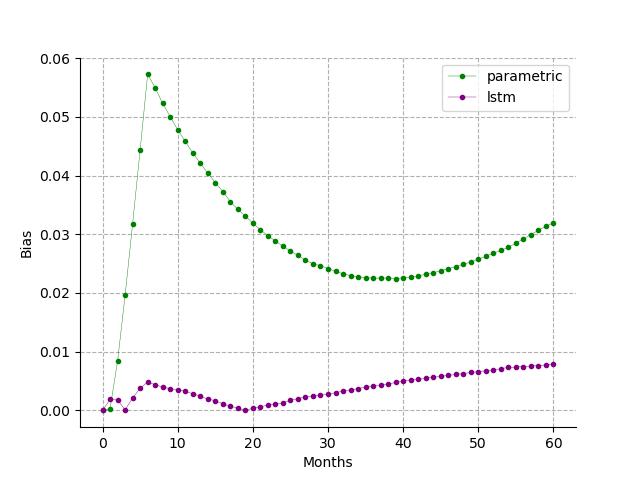} 
    \caption{Bias in risk under never treat}
    \label{fig:sfig16}
\end{subfigure}

\caption{Bias in risk in dataset with 10,000 individuals}
\end{figure}

\begin{figure}[p]
\centering
\begin{subfigure}{0.49\textwidth}
    \includegraphics[width=\linewidth]{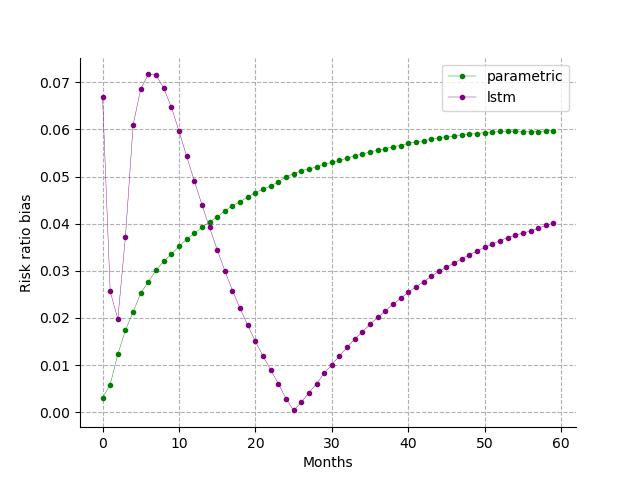} 
    \caption{Bias in risk ratio}
    \label{fig:sfig17}
\end{subfigure}
\hfill
\begin{subfigure}{0.49\textwidth}
    \includegraphics[width=\linewidth]{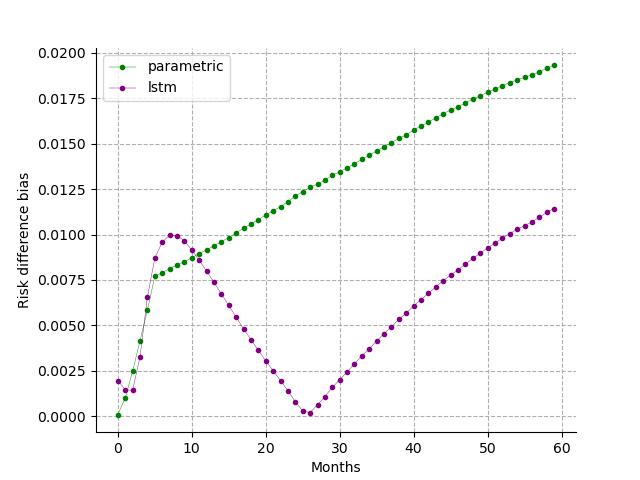} 
    \caption{Bias in risk difference}
    \label{fig:sfig18}
\end{subfigure}

\caption{Bias in effect estimates (always vs. never treat interventions) in 10,000 sample dataset}
\end{figure}

\FloatBarrier


\subsubsection{'Complex' time-dependency scenario}
\label{sec:risks_complex}

\begin{figure}[htbp]
\centering
\begin{subfigure}{0.49\textwidth}
    \includegraphics[width=\linewidth]{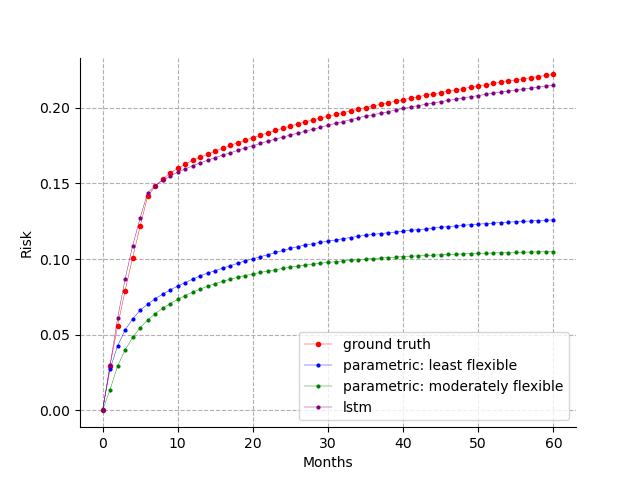} 
    \caption{Risk under always}
    \label{fig:sfig19}
\end{subfigure}
\hfill
\begin{subfigure}{0.49\textwidth}
    \includegraphics[width=\linewidth]{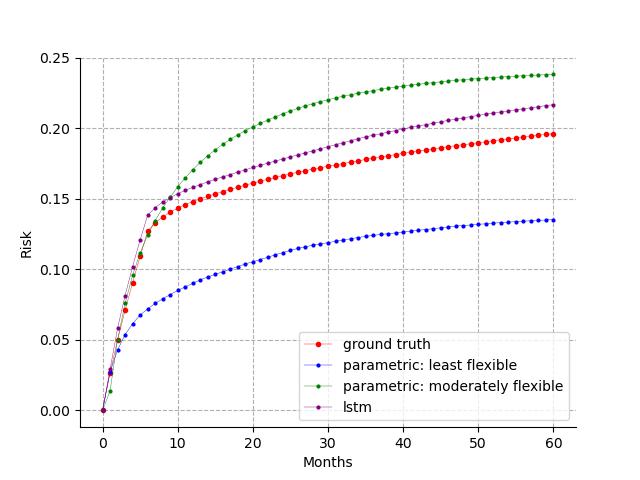} 
    \caption{Risk under never treat}
    \label{fig:sfig20}
\end{subfigure}
\caption{Ground truth and estimated risks under interventions in dataset with 1,000 individuals}
\end{figure}

\FloatBarrier 

\begin{figure}[htbp]
\centering
\begin{subfigure}{0.49\textwidth}
    \includegraphics[width=\linewidth]{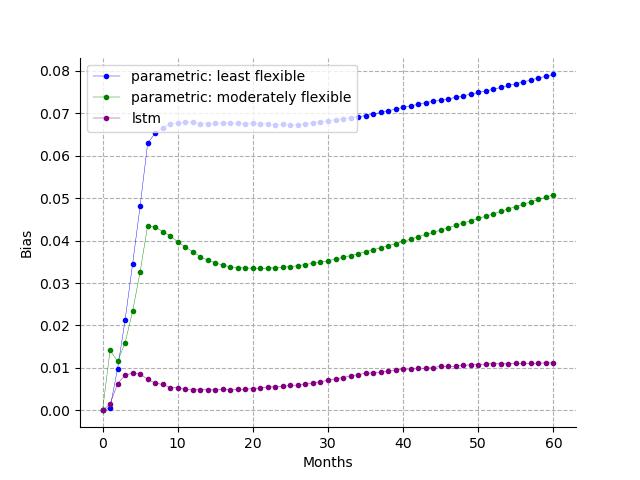} 
    \caption{Bias in risk under natural course}
    \label{fig:sfig21}
\end{subfigure}
\hfill
\begin{subfigure}{0.49\textwidth}
    \includegraphics[width=\linewidth]{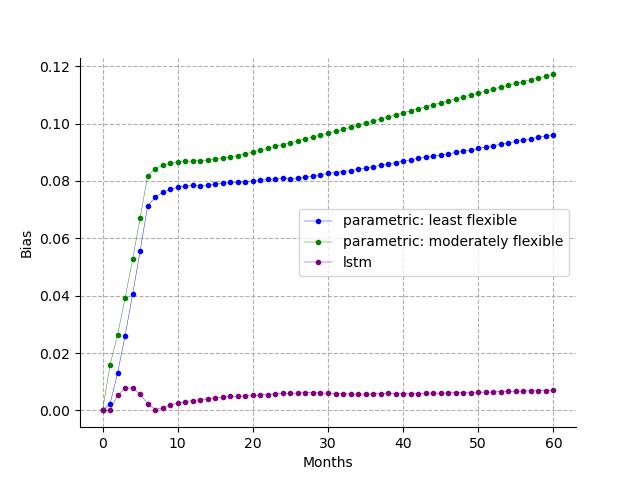} 
    \caption{Bias in risk under always treat}
    \label{fig:sfig22}
\end{subfigure}

\begin{subfigure}{0.49\textwidth} 
    \includegraphics[width=\linewidth]{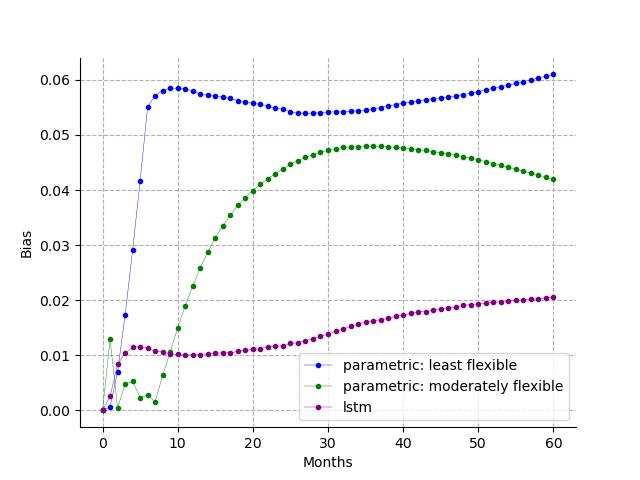} 
    \caption{Bias in risk under never treat}
    \label{fig:sfig23}
\end{subfigure}

\caption{Bias in risk in dataset with 1,000 individuals}
\end{figure}

\FloatBarrier 

\begin{figure}[htbp]
\centering
\begin{subfigure}{0.49\textwidth}
    \includegraphics[width=\linewidth]{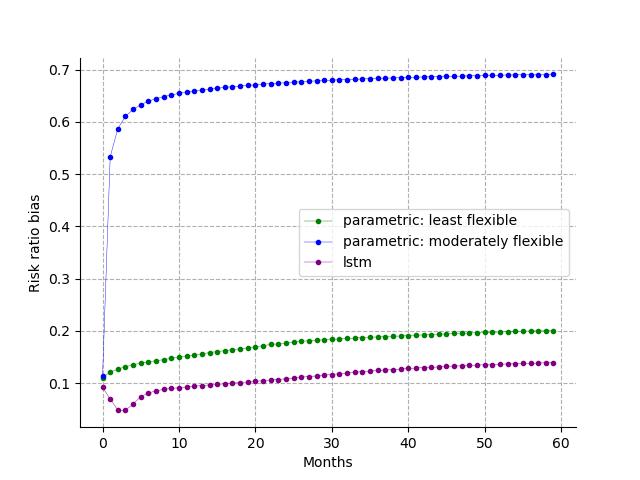} 
    \caption{Bias in risk ratio}
    \label{fig:sfig24}
\end{subfigure}
\hfill
\begin{subfigure}{0.49\textwidth}
    \includegraphics[width=\linewidth]{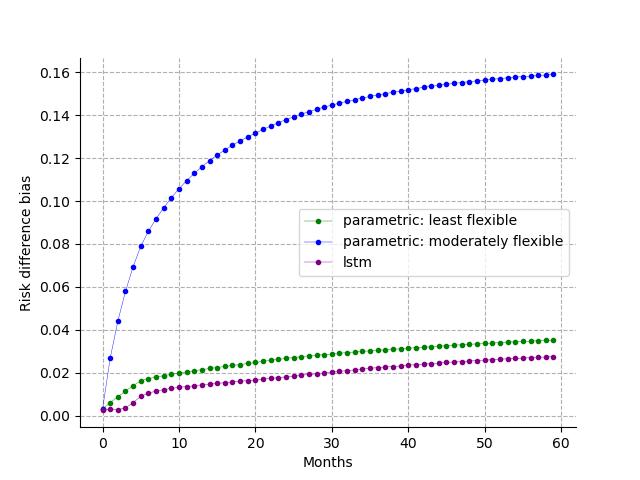} 
    \caption{Bias in risk difference}
    \label{fig:sfig25}
\end{subfigure}

\caption{Bias in effect estimates (always vs. never treat interventions) in 1,000 sample dataset}
\end{figure}

\FloatBarrier

\begin{figure}[p] 
\centering
\begin{subfigure}{0.49\textwidth}
    \includegraphics[width=\linewidth]{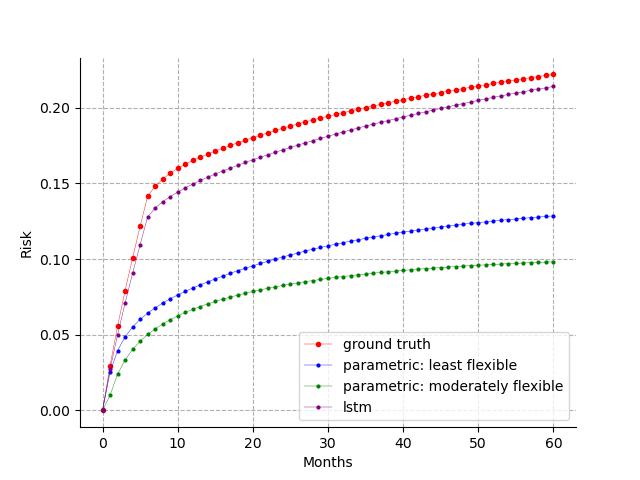} 
    \caption{Risk under always treat}
    \label{fig:sfig26}
\end{subfigure}
\hfill
\begin{subfigure}{0.49\textwidth}
    \includegraphics[width=\linewidth]{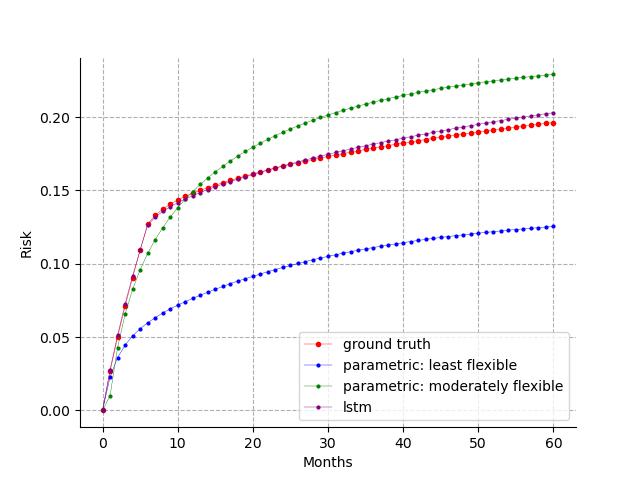} 
    \caption{Risk under never treat}
    \label{fig:sfig27}
\end{subfigure}
\caption{Ground truth and estimated risks under interventions in dataset with 10,000 individuals}
\end{figure}

\begin{figure}[p]
\centering
\begin{subfigure}{0.49\textwidth}
    \includegraphics[width=\linewidth]{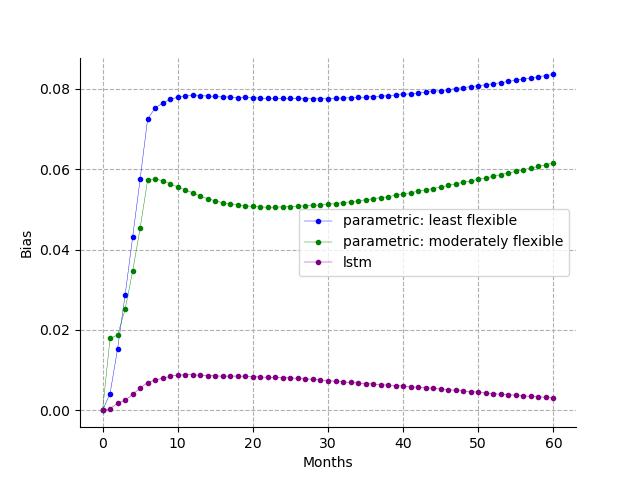} 
    \caption{Bias in risk under natural course}
    \label{fig:sfig28}
\end{subfigure}
\hfill
\begin{subfigure}{0.49\textwidth}
    \includegraphics[width=\linewidth]{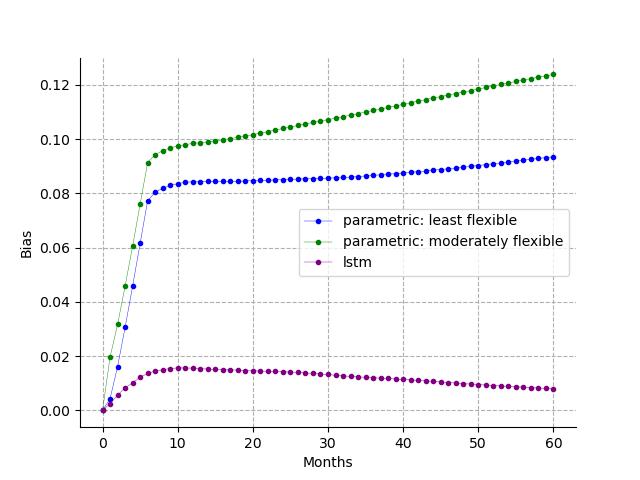} 
    \caption{Bias in risk under always treat}
    \label{fig:sfig29}
\end{subfigure}
\begin{subfigure}{0.49\textwidth} 
    \includegraphics[width=\linewidth]{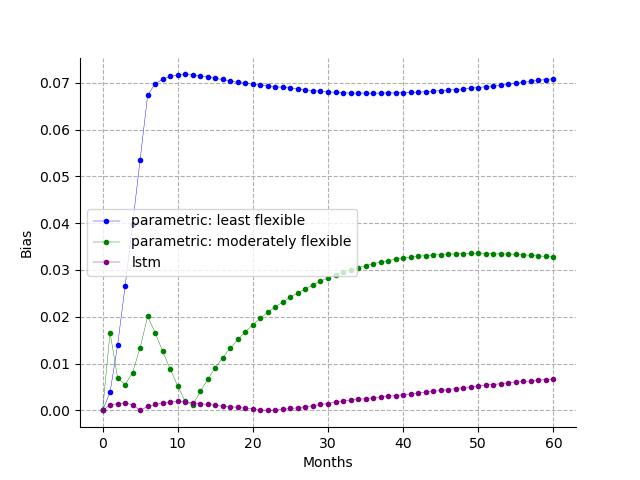} 
    \caption{Bias in risk under never treat}
    \label{fig:sfig30}
\end{subfigure}

\caption{Bias in risk in dataset with 10,000 individuals}
\end{figure}

\begin{figure}[p]
\centering
\begin{subfigure}{0.49\textwidth}
    \includegraphics[width=\linewidth]{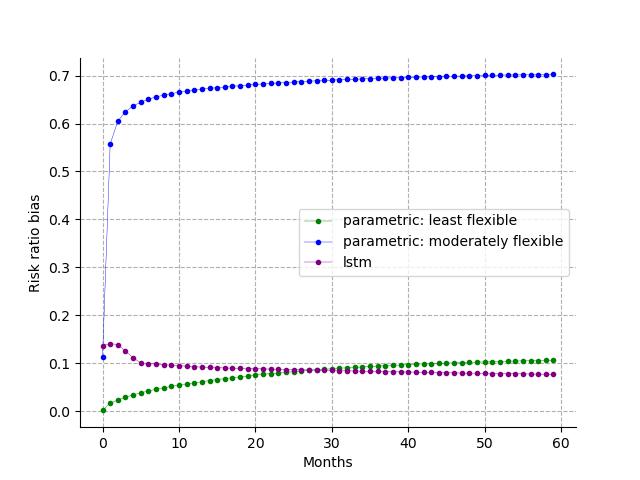} 
    \caption{Bias in risk ratio}
    \label{fig:sfig31}
\end{subfigure}
\hfill
\begin{subfigure}{0.49\textwidth}
    \includegraphics[width=\linewidth]{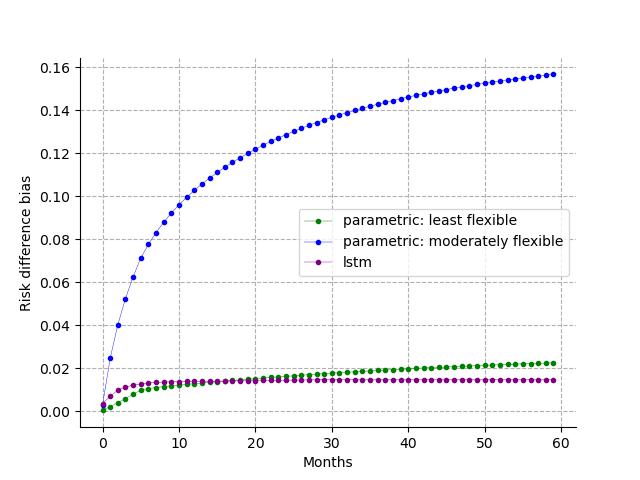} 
    \caption{Bias in risk difference}
    \label{fig:sfig32}
\end{subfigure}

\caption{Bias in effect estimates (always vs. never treat interventions) in 10,000 sample dataset}
\end{figure}

\FloatBarrier

\bibliographystyle{plain} 
\bibliography{Arxiv_Preprint_LSTM_only}


\end{document}